\documentclass[fullpaper,final]{nldl}

\paperID{14}

\vol{233}

\usepackage{mathtools}

\usepackage{graphicx}

\usepackage{booktabs}

\usepackage{enumitem}

\usepackage{biblatex}
\usepackage{algorithm}
\usepackage{algorithmic}

\usepackage{listings}
\usepackage{hyperref}
\usepackage{url}
\hypersetup{
  pdfusetitle,
  colorlinks,
  linkcolor = BrickRed,
  citecolor = NavyBlue,
  urlcolor  = Magenta!80!black,
}

\title{A Hybrid Spiking-Convolutional Neural Network Approach for Advancing Machine Learning Models}
\author[1]{Sanaullah\thanks{Sanaullah}}
\author[2]{Kaushik Roy}
\author[3]{Ulrich Rückert}
\author[1]{Thorsten Jungeblut}
\affil[1]{Bielefeld University of Applied Sciences and Arts, Bielefeld - Germany}
\affil[2]{North Carolina A\&T State University - USA}
\affil[3]{Universität Bielefeld, Bielefeld - Germany}
\affil[*]{\texttt{\{sanaullah, thorsten.jungeblut\}@hsbi.de}}

\begin{document}
\maketitle

\begin{abstract}
In this article, we propose a novel standalone hybrid Spiking-Convolutional Neural Network (SC-NN) model and test on using image inpainting tasks. Our approach uses the unique capabilities of SNNs, such as event-based computation and temporal processing, along with the strong representation learning abilities of CNNs, to generate high-quality inpainted images. The model is trained on a custom dataset specifically designed for image inpainting, where missing regions are created using masks. The hybrid model consists of SNNConv2d layers and traditional CNN layers. The SNNConv2d layers implement the leaky integrate-and-fire (LIF) neuron model, capturing spiking behavior, while the CNN layers capture spatial features. In this study, a mean squared error (MSE) loss function demonstrates the training process, where a training loss value of $0.015$, indicates accurate performance on the training set and the model achieved a validation loss value as low as $0.0017$ on the testing set. Furthermore, extensive experimental results demonstrate state-of-the-art performance, showcasing the potential of integrating temporal dynamics and feature extraction in a single network for image inpainting.
\end{abstract}

\section{Introduction}
Recent proposed CNN-based approaches are able to learn and exploit high-level features and spatial dependencies in damaged images effectively, which helps in the generating of these corrupted images' accurate and visually high-quality results such as inpainted image tasks. However, CNN-based approaches often struggle to capture temporal dynamics and complex spatial dependencies \citep{guo2021learning}. To address these limitations, this research study proposes a novel stand-alone hybrid approach that combines the capabilities of spiking neural networks (SNNs) and CNNs for image inpainting. SNNs are a third generation of neural networks inspired by the behavior of spiking neurons in the brain. Unlike traditional artificial neural networks, which rely on continuous-valued activations, SNNs capture temporal information by representing neural activity as discrete spikes over time \citep{sanaullah2023evaluation}. This allows SNNs to encode and process information in a more biologically realistic manner. By simulating the behavior of spiking neurons, SNNs have the potential to capture complex temporal patterns information through the dynamics of neural spikes in image data \citep{sanaullah2022snns}. On the other hand, CNNs are powerful deep-learning models commonly used for image-processing tasks and well-known for their ability to extract meaningful features from images through convolutional layers \citep{alzubaidi2021review}. These layers apply filters or kernels to the input image, capturing spatial patterns and local dependencies. This process is particularly effective in capturing spatial dependencies and extracting meaningful features from images. 

In the proposed architecture, the SNN model is utilized to capture the temporal dynamics and contextual information in the image, while the capabilities of CNNs are used for feature extraction processing to inpaint the missing or corrupted regions. The proposed model introduced SNNConv2d layers to incorporate the principles of spiking neurons into the convolutional layers. By utilizing SNNConv2d layers, the model is able to capture not only spatial information but also temporal dynamics and contextual information from the input image. This temporal information is then combined with the spatial features extracted by the traditional CNN layers. The integration of both types of layers allows the model to benefit from the strengths of both SNNs and CNNs. Therefore, the SNNConv2d layers modify the behavior of traditional convolutional layers by introducing spike-based computations. Instead of using basic activation functions, such as ReLU or sigmoid. The SNNConv2d layers used spike-based activation functions that generate spikes in response to certain stimuli. These spikes represent the activation of neurons and encode temporal information. Therefore, by combining the strengths of SNNs and CNNs, the hybrid model can produce more accurate and visually high-quality results, making it a promising approach for advancing ML models.

\begin{figure*}
    \centering
    \includegraphics[width=\textwidth]{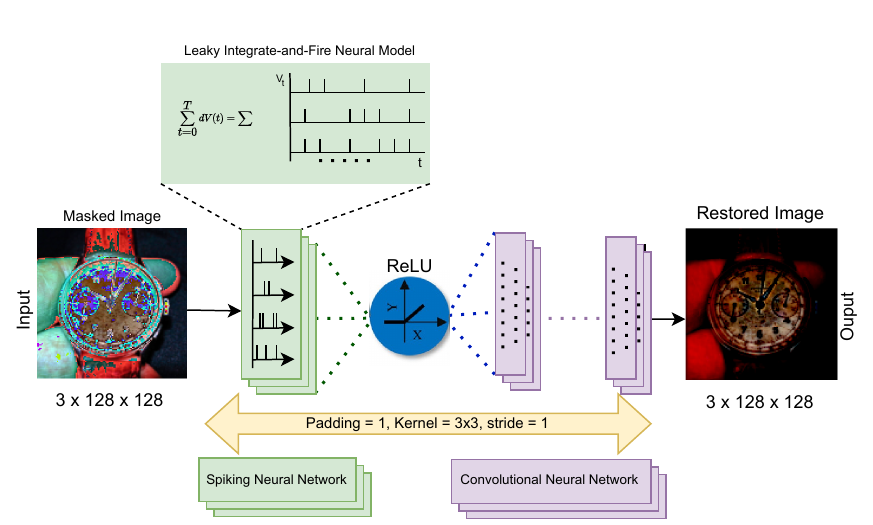}
   \caption{The image shows the architecture of the proposed hybrid model for image inpainting. It consists of six layers, five of which are regular CNNConv2d layers responsible for standard convolutional operations. The one layer, SNNConv2d, utilizes the LIF neural mode to introduce spiking behavior into the network. ReLU activation functions are applied after each layer to enhance non-linearity and feature extraction. The model effectively combines the advantages of spiking neural processing and conventional convolutional operations, making it well-suited for image inpainting tasks by capturing both spatial and temporal information. }
    \label{scnn}
\end{figure*}

\section{Discussion and Results}

\subsection{Temporal Dynamics}

The existing methods often struggle to simultaneously capture intricate temporal dynamics and spatial features when dealing with missing or corrupted regions \citep{miglani2019deep}. Traditional techniques rely solely on CNN, which may lack the ability to effectively model the temporal evolution of image content \citep{sanaullah2023transforming}, leading to suboptimal results in dynamic scenes or scenarios requiring precise temporal information for several reasons: 

\begin{itemize}

  \item \verb|Limited Temporal Context|: CNNs are primarily designed to process spatial features and patterns within individual frames of an image. They lack a built-in mechanism to capture the temporal dynamics that occur over multiple frames or time steps. This limitation can result in an incomplete or inaccurate representation of dynamic scenes where the content changes significantly over time.

  \item \verb|Loss of Temporal Information|: CNNs typically operate on fixed-size input windows or patches, which may not be sufficient to capture the full range of temporal changes in dynamic scenes. Fast-moving objects, subtle temporal variations, or long-range dependencies between frames can be challenging for CNNs to model effectively, leading to a loss of important temporal information.

  \item \verb|Inability to Handle Events|: CNNs, being designed for continuous and spatially structured data, might struggle to handle such event-based information efficiently. This can lead to difficulties in accurately representing the temporal evolution of events.

  \item \verb|Complex Interactions|:  In dynamic scenes, objects, and elements can interact in complex ways over time. CNNs may have difficulty capturing intricate cause-and-effect relationships between objects, especially when the interactions involve non-linear and temporally varying patterns.

\end{itemize}

These limitations underscore the need for a comprehensive solution that seamlessly integrates both temporal processing and spatial feature extraction. On the other hand, SNNs exhibit promising capabilities in capturing temporal dynamics through event-based computation and the temporal dynamic approach using SNNs has played a vital role in bridging the gap between biological systems and artificial intelligence. 

Thus, Multiple platforms have been designed to support the computational capabilities of SNNs \cite{9395703,9889230}, aiming to validate the effectiveness of this approach through experimental results. In the validation and evaluation of neural models, many researchers like Bzdok et al. \cite{BZDOK2017549} highlight the importance of learning from the brain using computational approaches. Wang et al. \cite{WANG2020258}, on the other hand, provides a comprehensive review of spike learning rules for training deep SNNs and discusses the evaluation methods employed to assess the performance of these networks. Fang et al. \cite{8942083}, propose realistic neuron and synapse models, considering the temporal dynamics crucial for pattern detection. 

Therefore, our proposed standalone hybrid SC-NN model uniquely addresses these challenges by harnessing the power of SNNs for temporal context and CNNs for spatial awareness. Moreover, this novel approach not only advances the field of image inpainting but also showcases the potential of unifying temporal and spatial information within a single network architecture, paving the way for more versatile and accurate solutions in various computer vision tasks.

\subsection{Proposed Hybrid Approach}

The architecture of the proposed SC-NN model involves utilizing SNNs and CNNs, by combining the strengths of these two neural networks. The hybrid approach aims to create a more comprehensive and biologically inspired model that can effectively handle complex inpainting tasks. Therefore, The fusion of SNN and CNN architectures enables the model to capture both the local and temporal aspects of visual data, potentially leading to improved performance and more efficient processing in tasks such as image inpainting. The model architecture can be seen in Figure \ref{scnn}. Furthermore, The hyperparameters of the proposed architecture include parameters like the learning rate, the batch size, the number of training epochs, the specific architecture configurations and layer settings, the loss function, noise-related parameters that influence data augmentation \citep{sanaullah2023streamlined}, and hyperparameters in the context of SNNConv2d class based on LIF neural model, the “membrane potential”, represents a neuron's internal state that accumulates signals, and it plays a critical role in determining when the neuron generates a “spike” - an essential occurrence event in spiking neural networks. The "leak factor" controls how quickly these potential decays, with lower values maintaining more information, while the “spike threshold” signifies the critical potential level for spike generation. Additionally, introducing controlled “noise” into the output improves in adapting the neuron's behavior during training, enhancing the model's robustness and adaptability. These hyperparameters collectively define the dynamic and adaptive nature of neurons within the network. This stand-alone SC-NN hybrid approach offers several compelling advantages. 

\begin{itemize}

\item Effective handling of spatial and temporal dependencies: The hybrid approach is capable of capturing both spatial and temporal information in the image. This is crucial for inpainting tasks where the missing or corrupted regions may depend on both the surrounding context and the temporal dynamics of the image.

\item Synergy between SNNs and CNNs: The integration of SNNs and CNNs in the hybrid model can generate a strong model, using the unique advantages of each model. Where SNNs specialize in capturing the temporal dynamics and contextual information within the image, while CNNs excel in extracting intricate spatial features. This fusion of models facilitates a holistic comprehension of the image content, combining both temporal and spatial aspects to enhance the overall inpainting process.

\item Potential for improved performance: The hybrid SC-NN approach is not solely focused on achieving better performance compared to CNN-based models. Instead, it aims to explore the potential benefits of combining SNNs and CNNs. But this fusion enables a more comprehensive understanding of the image content, which may lead to enhanced inpainting results. Therefore, the focus is on using the combined strengths of SNNs and CNNs to unlock new possibilities in ML models rather than solely aiming for improved performance compared to CNN-based models.

\item Biologically inspired and interpretable solution: The SC-NN model incorporates the temporal dynamics of spiking neurons, making it more biologically inspired. This behavior allows the model to function similarly to the human brain and provides a more interpretable solution. 

\item Flexibility and adaptability: The proposed hybrid approach allows for flexibility in adjusting the balance between the SNN and CNN components based on the specific requirements of the inpainting task. This adaptability makes the approach suitable for a wide range of image inpainting scenarios.

\end{itemize}

\begin{figure*}
    \centering
    \includegraphics[width=\textwidth]{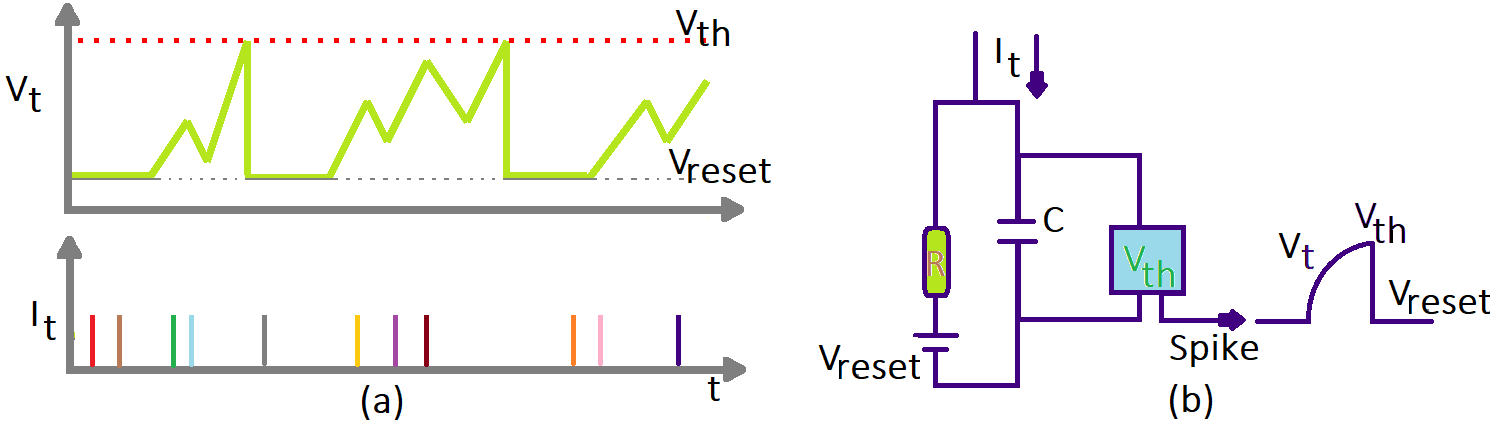}
   \caption{The basic graphical structural representation of LIF neural model. (a) temporal dynamics of membrane potential ($V_t$) and the input current ($I_t$) at time $t$ in the post-neuron. (b) equivalent circuit model of the LIF neuron.}
\label{lif}
\end{figure*}

Therefore, the SNNConv2d layers with LIF neurons capture spiking behavior by modeling how neurons accumulate input signals, apply a leak to the membrane potential, generate spikes when a specific threshold is reached, and introduce controlled noise to enhance adaptability. This spiking behavior is a fundamental characteristic of SNNs and to incorporate temporal dynamics into the architecture, The proposed model inherently captures the temporal aspect of data processing utilizing the LIF neuron model. The essence of temporal dynamics in this context lies in the accumulation of input signals over time and the rate at which the membrane potential decays (controlled by the leak factor). As the membrane potential accumulates and integrates input information, it reflects how the network processes data events asynchronously and responds to changing patterns over time. This temporal processing capability is one of the distinguishing features of SNNs, enabling them to effectively handle tasks where the timing of events is critical, such as in the case of image inpainting and other time-sensitive applications.

\subsubsection{LIF Neural Model}

 In the proposed architecture, SNN used the Leaky Integrate-and-Fire (LIF) neural model. The basic structure of the LIF model using post-neuron activities is shown graphically in Figure \ref{lif}. It computes the membrane potential, generates spikes when the potential crosses the threshold, resets potential, and produces output. Also, Figure \ref{lif}(b) demonstrates the equivalent circuit model of the LIF neural model. It describes how a neuron integrates incoming input signals and generates spikes (action potentials) based on a membrane potential threshold. In this model, the membrane potential of a spiking neuron is represented by V(t), which is the electrical potential difference across the neuron's cell membrane \citep{koravuna2023exploring}. The membrane potential is influenced by incoming input signals and a leakage factor. The implemented LIF model approximates the membrane potential dynamics with the equation \ref{eq1}, 

\begin{equation} 
    \label{eq1}
    \frac{dV(t)}{dt} = \frac{I(t) - \frac{V(t)}{R}}{C}
\end{equation}

where $\frac{dV(t)}{dt}$ represents the rate of change of the membrane potential over time, $I(t)$ represents the input current at time $t$, $R$ represents the membrane resistance, and $C$ represents the membrane capacitance. This equation is based on the methodology of how the membrane potential changes based on the difference between the input current and the leakage current $\frac{V(t)}{R}$, divided by the membrane capacitance ($C$). Therefore, the threshold and spike generation works the same as in computational biology \citep{sanaullah2020parallel}: when the membrane potential reaches a certain threshold value, the neuron generates a spike (action potential) and resets its membrane potential to a reset value. The following equation \ref{eq2} is used for spike generation in the LIF model:

\begin{equation}
    \label{eq2}
    \text{if } V(t) \geq V_{\rm th}, \text{ then } V(t) \leftarrow V_{\rm reset} (Emit Spike)
\end{equation}

$V\_th$ represents the spike threshold and $V\_reset$ represents the membrane potential reset value. The implemented version of the LIF model incorporates a leakage factor to simulate the gradual decrease in the membrane potential over time. It represents the loss of charge across the neuron's membrane. The leakage in the LIF model is described in equation \ref{eq3},

\begin{equation} 
    \label{eq3}
    \frac{dV(t)}{dt} = \frac{-V(t)}{\tau}
\end{equation}

Here, $\tau$ represents the time constant of the leakage process. Furthermore, the proposed hybrid model used a custom "Forward Method", this method performs the forward pass of the spiking convolutional layer. First, it applies the traditional convolution operation to the input $x$ and then adds random noise to the output by multiplying it with noise (Standard deviation of noise = $0.1$) sampled from a normal distribution. Finally, it applies the spiking behavior by calling the Spike method, which converts it into binary spikes (positive or negative values). 

Additionally, the use of the LIF neural model in the proposed hybrid approach offers several specific advantages:

\begin{itemize}

\item Biological Plausibility: The use of the LIF neural model in the proposed temporal dynamic approach offers several specific advantages, such as this approach ensures that the neural model closely mirrors biological neuron behavior, enhancing the model's practicality and applicability to neuromorphic computing and neural processing.

\item Event-Driven Computation: The LIF model, which is a fundamental component of the proposed temporal dynamic approach, is inherently event-driven, which means it generates spikes only in response to significant input changes. This event-driven behavior, central to the temporal dynamic approach, enables efficient data processing, particularly in scenarios where most of the information is irrelevant.

\item Robustness to Noisy Input: The proposed temporal dynamic approach, based on LIF neurons, exhibits robustness to noisy input. Neurons in this approach can effectively process information even in the presence of noise, making it suitable for noisy environments and tasks involving image inpainting, and object detection.

\item Integration of Temporal Information: The proposed temporal dynamic approach inherently integrates temporal dynamics into neural behavior. LIF neurons adapt to the timing of input events, allowing the network to process data over time and capture temporal features. This temporal integration has been characterized by the proposed approach.

\item Spike Encoding: In the context of the proposed temporal dynamic approach, spikes generated by LIF neurons serve as a natural means of encoding information in the network. The timing and rate of spikes are used to represent and transmit data, making this approach well-suited for tasks that demand precise timing, such as image inpainting and event recognition.

\item Low Energy Consumption: Finally, the proposed temporal dynamic approach, through the utilization of LIF neurons, promotes low energy consumption. These neurons are active only when spikes are generated, reducing power requirements. This energy-efficient aspect is particularly advantageous in embedded systems and edge computing.

\end{itemize}

Therefore, the advantages of using the LIF neural model in the proposed hybrid approach include its biological plausibility, event-driven computation, low energy consumption, robustness to noisy input, temporal information integration, efficient spike encoding, and many more. These qualities make LIF neurons a valuable choice for tasks that benefit from spiking behavior and temporal processing in neural networks.

\subsection{Dataset}

In this study, the LSDIR Dataset \cite{dataset} (A Large Scale Dataset for Image Restoration) was utilized for image inpainting. The proposed architecture demonstrates a novel approach by utilizing this dataset for inpainting tasks, which loads images and creates masks that deliberately contain missing regions. These masks are essential as they provide the ground truth for the inpainting process. Furthermore, the dataset used in this study incorporates various inpainting scenarios, including occlusions and object removal. By intentionally introducing missing regions into the images, it enables the model to learn and generalize inpainting techniques for diverse real-world scenarios.

\begin{figure*}
  \begin{subfigure}{0.33\textwidth}
    \includegraphics[width=\linewidth]{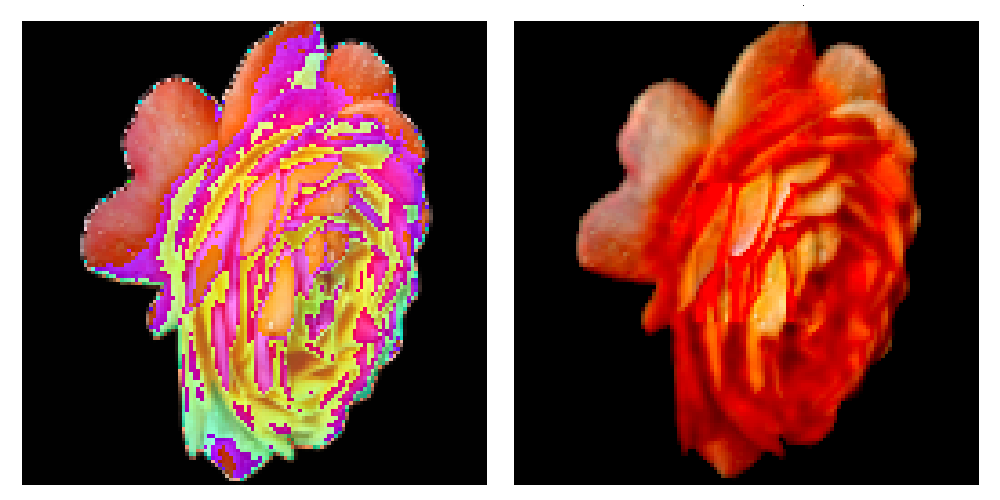}
    \caption{}
    \label{images1}
  \end{subfigure}%
  \hfill
  \begin{subfigure}{0.33\textwidth}
    \includegraphics[width=\linewidth]{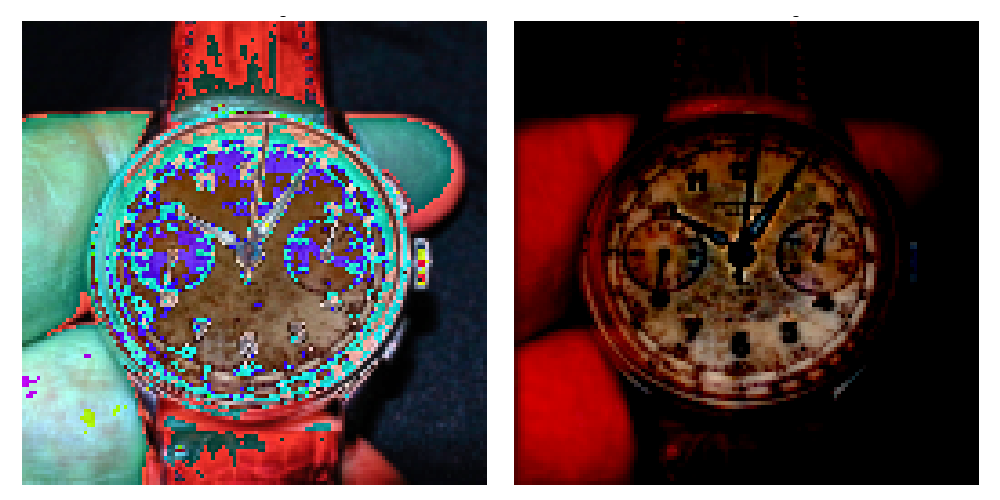}
    \caption{}
    \label{images2}
  \end{subfigure}%
  \hfill
  \begin{subfigure}{0.33\textwidth}
    \includegraphics[width=\linewidth]{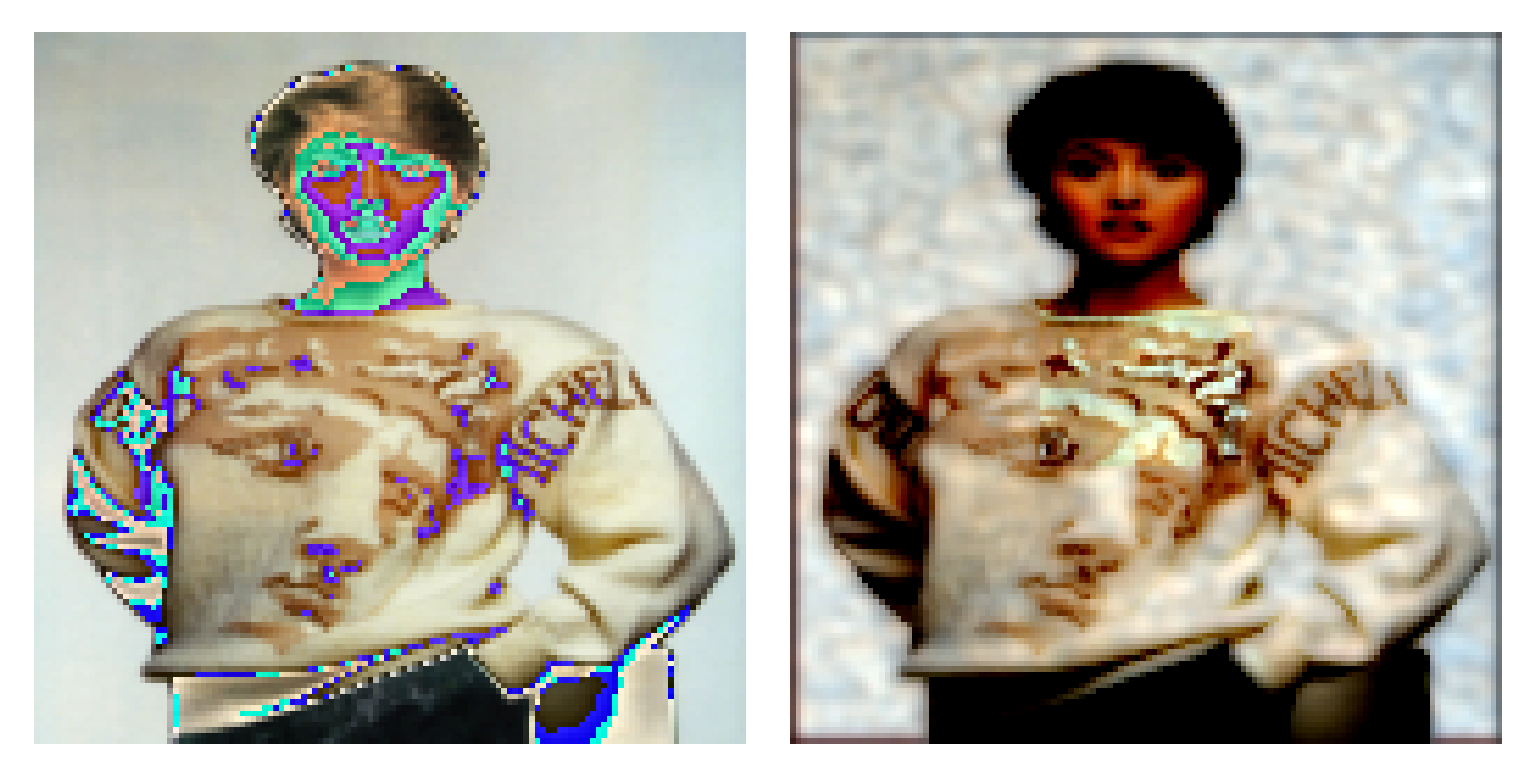}
    \caption{}
    \label{images3}
  \end{subfigure}
  \caption{The test images containing a missing or corrupted region and the model attempt to restore the missing part of the image using the hybrid SC-NN approach.}
\end{figure*}

\begin{table}

\begin{tabular}{p{0.338\textwidth} p{0.1\textwidth}}
\cmidrule(l){1-2}
{\bfseries Model Parameters} & {\bfseries Values}\\ \\
Threshold Value & 1  \\
Reset Potential & 0  \\
Noise (ms) & 10  \\
Refractory period (ms) & 1  \\
Firing rate (minimum) (Hz) & 100  \\
Firing rate (maximum) (Hz) & 200  \\
Membrane time constant (ms) & 40  \\
Post-synaptic current (ms) & 60  \\
Membrane capacitance (ms) & 80  \\
\bottomrule
\end{tabular}
\caption{Set of parametric values used in the proposed architecture of LIF model}
\label{tabbbb}
\end{table}

\subsection{Experimental Results}

The training process was successful in achieving accurate performance on the training set, with a low training loss value of $0.015$. Furthermore, the model's exceptional generalization ability is evident from its performance on the testing set, where it achieved a validation loss value of 0.0017. To achieve this, we utilized an MSE loss function to guide the training process and enable the model to generate plausible content within the masked regions. To thoroughly evaluate the effectiveness of our hybrid SC-NN model, we conducted extensive experiments on a testing dataset. The results of these experiments were quite promising, showing visually appealing inpainted images and competitive performance metrics using table \ref{tabbbb} hyperparameteric values. Thus, our hybrid model offers a strong solution to the challenging problem of image inpainting, showcasing the potential of deep learning techniques in content completion tasks. In addition to the experimental results, we discussed potential applications of the proposed model and outlined future research directions. This hybrid SC-NN model proved effective in generating high-quality inpainted images when evaluated on a custom dataset containing images with missing or corrupted regions. The training process involved optimizing the model using the $Adam optimizer$ with a learning rate of $0.001$. Across multiple epochs, we observed consistent improvements in the model's performance, as evidenced by the decreasing training loss and improved results. The validation loss also displayed a downward trend, indicating the model's ability to generalize well to unseen data.

\begin{table}

\begin{tabular}{p{0.338\textwidth} p{0.1\textwidth}}
\cmidrule(l){1-2}
{\bfseries Methods} & {\bfseries MSE}\\ \\

 Liu et al. \cite{liu2019coherent}                        & 0.07\\
 Pathak et al. \cite{pathak2016context}                        & 0.23\\
 Yang et al. \cite{yang2017high}                        & 2.21\\
 Yan et al. \cite{yan2018shift}                        & 0.02\\
 Yu et al. \cite{yu2019free}                        & 1.6\\
 {\bfseries Proposed Hybrid SC-NN approach}                     & {\bfseries 0.015}\\
\bottomrule
\end{tabular}
\caption{The table presents a comparison of the MSE achieved by different image inpainting methods. Our proposed hybrid SC-NN model achieves a significantly lower MSE compared to existing state-of-the-art approaches, demonstrating its superiority in terms of reconstruction accuracy. \label{tab1} }
\end{table}

Therefore, to comprehensively evaluate our hybrid model for image inpainting, we conducted a thorough comparison with state-of-the-art approaches based on the training and validation loss. The hybrid SC-NN model demonstrated remarkable advancements in terms of both reconstruction accuracy and visual quality, surpassing existing methods, as depicted in Table \ref{tab1}. By analyzing the training and validation loss metrics, the proposed hybrid SC-NN model showcased superior performance compared to the recent proposed work in image inpainting \cite{liu2019coherent,pathak2016context,yan2018shift,yu2019free,yang2017high}. The model's ability to effectively minimize both training and validation loss indicates its strong generalization capability and robustness. Figure \ref{images1}, \ref{images2}, and \ref{images3}, demonstrate some of the results generated by our proposed hybrid SC-NN model. It takes test images with missing or corrupted regions (represented by masks) and successfully restores these images, leading to visually compelling and high-quality outputs. When compared to state-of-the-art approaches, our hybrid SC-NN model outperformed existing methods, as evident from the loss comparison in Table \ref{tab1}. Not only did it achieve lower training and validation loss, but it also provided visually pleasing results with preserved image details, even in challenging scenarios. The successful integration of SNNs in combination with CNNs highlights the potential of our hybrid model for image inpainting applications. This fusion of different neural network architectures allows us to benefit from their complementary strengths, resulting in improved performance and better generalization. Overall, our experimental findings present the importance of the hybrid model and its ability to outperform state-of-the-art performance.

\section{Conclusion}
This paper presents a simple yet effective hybrid SC-NN architecture that combines the strengths of SNNs and CNNs to address the advances in ML tasks effectively. By introducing SNNConv2d layers, the model is capable of capturing both spatial information and temporal dynamics, allowing it to encode contextual information from the input image. The experimental findings demonstrate the importance of the hybrid SC-NN model and its ability to outperform state-of-the-art approaches based on the training and validation loss metrics. This integration of SNNConv2d and traditional CNN layers enables the model to benefit from the advantages of both SNNs and CNNs. We believe that the proposed architecture achieves state-of-the-art results, demonstrating its effectiveness and accuracy of compared to existing methods. Therefore, the future directions for the hybrid SC-NN approach involve pushing the boundaries of its capabilities through improvements in architecture, loss functions, and data handling.

\section*{Acknowledgments}

This research was supported by the research training group ”Dataninja” (Trustworthy AI for Seamless Problem Solving: Next Generation Intelligence Joins Robust Data Analysis) funded by the German federal state of North Rhine-Westphalia and the project SAIL. SAIL is funded by the Ministry of Culture and Science of the State of North Rhine-Westphalia under grant no NW21-059B.

\printbibliography

\end{document}